\documentclass[runningheads]{llncs}
\usepackage{eccv}

\usepackage{eccvabbrv}

\usepackage{graphicx} 
\usepackage{booktabs}
\usepackage{multirow} 
\usepackage{array} 
\usepackage{makecell} 
\usepackage[table]{xcolor} 
\usepackage{amsmath} 
\usepackage{amssymb} 
\usepackage{enumitem} 
\usepackage{microtype} 
\usepackage{pifont} 
\usepackage{xspace} 
\usepackage{wrapfig}
\usepackage{url}
\usepackage{placeins}
\usepackage{marvosym}

\usepackage{overpic} 
\usepackage{caption} 
\usepackage{subcaption} 
\usepackage[accsupp]{axessibility}  
\usepackage[acronym]{glossaries}

\usepackage[pagebackref,breaklinks,colorlinks,citecolor=eccvblue]{hyperref}

\usepackage[capitalize]{cleveref} 

\usepackage{orcidlink}

\definecolor{mygray}{rgb}{0.9,0.9,0.9} 
\definecolor{novelcolor}{rgb}{0.8,1.0,0.9} 

\crefname{section}{Sec.}{Secs.} 
\Crefname{section}{Section}{Sections} 
\Crefname{table}{Table}{Tables} 
\Crefname{Figure}{Figure}{Figures} 

\makeglossaries 
\newacronym{XAI}{XAI}{Explainable AI} 
\newacronym{LLM}{LLM}{Large Language Models} 
\newacronym{VLMs}{VLMs}{Vision--Language Models} 
\newacronym{MCQs}{MCQs}{Multiple-Choice Questions} 
\newacronym{VQA}{VQA}{Visual Question Answering} 
\newacronym{DL}{DL}{Deep Learning} 
\newacronym{AI}{AI}{Artificial Intelligence} 
\newacronym{CDDM}{CDDM}{Crop Disease Domain Multimodal Dataset} 
\newacronym{TomaMMU}{TomaMMU}{Tomato leaf disease MultiModal Understanding}
\newacronym{IoT}{IoT}{Internet of Things}
\newacronym{OEQs}{OEQs}{Open-Ended Questions}

\begin{document}

\title{TomaMMU: A Comprehensive Multimodal Understanding Benchmark for Tomato Leaf Diseases} 
\titlerunning{TomaMMU Benchmark for Tomato Leaf Diseases} 

\author{
    Gia-Han Truong\inst{1}\thanks{Equal contribution}\orcidlink{0009-0000-3286-3212} \and 
    Khang Nguyen Quoc\inst{2}\textsuperscript{\text{$\star$}}\orcidlink{0000-0003-4927-4822} \and 
    Luyl-Da Quach\inst{1}\textsuperscript{\Letter}\orcidlink{0000-0002-5661-4250}
}
\authorrunning{Truong et al.} 

\institute{
    Department of Information Technology, FPT University, Can Tho, Vietnam \and 
    School of Electrical Engineering, Korea University, Seoul, South Korea \\ 
    \email{truonggiahan0702ct@gmail.com, khangnq@korea.ac.kr, luyldaquach@gmail.com}
}  

\maketitle
\begin{abstract}
Despite advances in automated plant disease recognition, existing systems primarily frame diagnosis as a visual classification problem, leaving multimodal disease understanding and reasoning underexplored, particularly in real-world settings marked by visual variability, background clutter, and inconsistent illumination. To address this gap, we introduce TomaMMU, a large-scale \textbf{Toma}to leaf disease \textbf{M}ulti\textbf{M}odal \textbf{U}nderstanding dataset, alongside TomaBench, a benchmark for evaluating  \gls{VLMs} on tomato disease understanding.
TomaMMU comprises 28,808 high-quality images spanning 15 categories and 213,119 human-annotated visual question-answer pairs, generated through a three-stage pipeline comprising Data Collection, Human Annotation, and Question-Answer Generation. 
Building on this foundation, TomaBench organizes seven agricultural tasks into a hierarchical three-level taxonomy spanning Basic Perception, Pathology Understanding, and Expert Diagnosis, which together enable systematic evaluation from low-level visual recognition to high-level diagnostic reasoning. The tasks assess visual symptom recognition, taxonomic relationships, and diagnostic reasoning, offering a comprehensive view of how well models grasp plant pathology. Our results pronounced gaps in fine-grained recognition and factually grounded reasoning with 14 state-of-the-art VLMs, consistently underperforming on both challenging \gls{MCQs} and open-ended questions.
These results suggest that current \gls{VLMs} struggle to translate visual perception into reliable diagnostic knowledge, motivating the need for targeted domain adaptation. Simple fine-tuning on TomaMMU substantially narrows this gap, boosting accuracy on challenging \gls{MCQs} to 96.09\%, outperforming recent VLMs, and pointing toward promising directions for future work. All data and code is available in \url{https://huggingface.co/datasets/enalis/TomaMMU}

\end{abstract}

\section{Introduction}
\label{sec:intro}
Multimodal understanding and reasoning are crucial for plant disease identification, helping the system overcome environmental influences in decision-making and also providing farmers and experts with a more intuitive understanding of plant diseases. Survey studies have shown that many studies have used various machine learning methods, from conventional vision models to advanced \gls{VLMs}, following a process of data collection, preprocessing, segmentation, feature extraction, classification, and evaluation, but are limited by environmental factors, interpretability, lack of field data, and especially generalization (where a model performs well on a tomato dataset but not on other data types) \cite{Jafar2024, Khan2025, Sajitha2024}. Hence, this study proposes a new process for building a large-scale, standardized dataset to provide detailed assessments of plant diseases, support understanding, and enable high-level inferences, including assessments of tomatoes, thereby enabling further research to be expanded and developed in the future.

Recent advances in smart agricultural adaptation, such as  \gls{AI}, the \gls{IoT}, and \gls{DL}, are achieving promising results in addressing the large volume of data collected from cameras, satellites, and drones. George et al. surveyed many studies that applied \gls{IoT}, \gls{XAI}, Convolutional Neural Networks (CNNs), and Vision Transformers (ViTs) to benchmark datasets (PlantVillage \cite{PlantVillage}). However, all encountered difficulties in real-world conditions \cite{George2025}, with accuracy typically dropping by 30-40\% when the model was trained on a laboratory dataset (PlantDoc \cite{PlantDoc}, FieldPlant \cite{FieldPlant}). To clarify model-dependent classification, studies have been conducted using \gls{XAI} to explain the dependence between feature-related outcomes and advanced \gls{DL} models to address dependence on field outcomes such as background removal, focusing on CNN model changes, but also depending on real-world images, background clutter, occlusions, leaf pose variation, and natural symptom diversity still impacting the results \cite{Quach2025, SenthilPandi2022}. These factors, such as lighting, background clutter, occlusions, leaf pose variation, and natural symptom diversity, reveal that many models rely on superficial correlations rather than robust disease representations, leading to poor generalization in field environments. The stakes are also high in this scenario, as it directly impacts food security and farmer livelihoods. This performance degradation is due to the lack of comprehensive benchmarks to evaluate and improve \gls{VLMs}' ability to understand agricultural problems, especially to address real farmers and experts concerns. 

Creating such a benchmark presents various multifaceted challenges. First, biological and agricultural tasks, which often use computer vision, are notably labor-intensive for data collection (labeling is a first-order requirement) \cite{Khan2025}. However, such domain knowledge is scarce and highly specialized, making it exceedingly difficult to curate high-quality evaluation datasets. Second, to address environmental variability, robust systems must go beyond simple image classification by combining the visual observation with background information \cite{Zhu2025}. Third, no clear protocol for defining a representative distribution of realistic agricultural questions \cite{Tzachor2023}. These challenges have created a significant gap in understanding whether \gls{AI} systems can handle real-world agricultural problems.

\begin{figure}[t]
\centering
\includegraphics[
    width=\linewidth
]{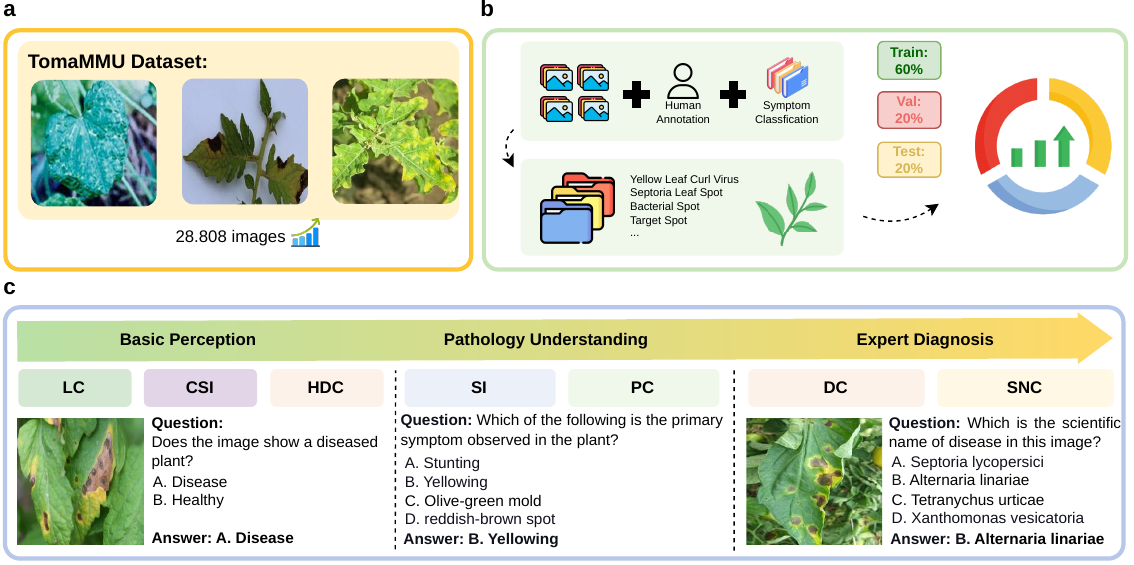}
\caption{The \gls{TomaMMU} Curation and Benchmarking Pipeline. \textbf{a,} Raw image data of tomato leaf types were collected from in-the-wild sources and in-house collections. \textbf{b,} Humans participated in manual labeling and classification of disease symptoms. A metadata set with Train/Val/Test sets proportionally divided into 60\% / 20\% / 20\% portions. \textbf{c,} Question and Answer Dataset. From metadata-image pairs, we developer TomaBench, a curated subset designed to benchmark Large Vision-Language Models on image classification, zero-shot learning, and \gls{VQA}.}
\captionsetup[figure]{name=Figure}
\label{fig:data}
\end{figure}

To address the aforementioned challenges, we introduce the Tomato leaf disease MultiModal Understanding \gls{TomaMMU} dataset, designed to support the development of \gls{VLMs} for research on tomato leaf disease. Unlike other datasets that rely on generating images with a uniform laboratory background and synthetic annotations via \gls{AI}, \gls{TomaMMU} integrates curated human-label annotations. It also includes images collected from other authoritative sources, such as LeafNet \cite{LeafNet}, enabling evaluation under real-world conditions rather than solely recognizing surface patterns in a controlled environment. For data curation, we design a pipeline to extract tomato leaf knowledge from long-form expert answers, and then have human annotators evaluate the quality and finalize the question-answer pairs. Based on the real-world questions we collected, the final dataset covers seven major types of tomato questions and knowledge: crop species identification, healthy-diseased classification, single-many leaves classification, disease classification, symptom identification, pathogen classification, and scientific name classification, as shown in Fig. \ref{fig:data}. Including with \gls{TomaMMU} is TomaBench, an assessment framework with 42,626 question-answer pairs covering 3 hierarchical diagnostic tasks (from basic health screening to in-depth scientific name classification), a comprehensive evaluation resource designed to assess the progress of \gls{VLMs} in understanding tomato pathology. The proposed dataset and evaluation framework are expected to significantly accelerate research progress in understanding multimodal tomato disease.

Overall, the main contributions of this study are as follows:
\begin{itemize}
    \item We provide \gls{TomaMMU}, a \gls{VQA} dataset for tomato leaf disease comprising more than 28,808 images and up to 213,119 \gls{VQA} pairs across 14 tomato diseases and 1 tomato healthy, along with seven \gls{VQA} tasks, which are Healthy Diseases Classification (HDC), Disease Classification (DC), Pathogen Classification (PC),  Crop species identification (CSI), Scientific Name Classification (SNC), Leaf Count (LC) and Symptoms Identification (SI). These tasks test the extent to which \gls{VLMs} can support disease diagnosis by integrating image information into real-world environments.
    \item Introducing TomaBench, a comprehensive benchmarking framework for tomato disease detection and characterization. 
    \item A systematic evaluation of the leading \gls{VLMs} reveals their limitations in handling knowledge-intensive queries about tomato diseases, along with an error analysis. 
\end{itemize}

\section{Related Work}\label{relatedwork}


\noindent \textbf{Multimodal Datasets and Benchmarks}. General-domain multimodal benchmarks have evolved from basic visual question answering toward complex, real-world reasoning. These datasets, such as \gls{VQA}v2~\cite{VQAv2} and GQA~\cite{GQA} established the foundation for visual question answering, while ScienceQA~\cite{ScienceQA} introduced scientific reasoning, though its academic-source samples limit real-world relevance. More recent benchmarks push further into complex reasoning and real-world grounding: RealWorldQA~\cite{yue2024mmmu} targets everyday visual queries, and Eyes-Wide-Shut probes whether models rely on genuine visual grounding rather than language priors~\cite{tong2024eyes}. Within the agriculture domain, multimodal datasets remain comparatively scarce and narrow in scope, often sacrificing scale, realism, or reasoning depth. Larger-scale efforts have since emerged to address the coverage gap, such as the \gls{CDDM}~\cite{cddm} dataset, which has expanded to 137,000 images spanning 16 crop species and 60 disease categories, paired with nearly one million question-answer pairs. Beyond scale, other benchmarks push toward richer reasoning: AgriBench~\cite{AgriBench}, and LeafBench~\cite{LeafNet} organizes agricultural questions into a hierarchical structure that evaluates both surface-level understanding and deeper reasoning designed for vision-language understanding of plant diseases. 

Even with these advances, however, no existing agricultural dataset combines large-scale, field-collected imagery with the depth of hierarchical reasoning needed to fully capture real-world diagnostic complexity, a gap that \gls{TomaMMU} is designed to close.

\noindent \textbf{Vision-language models for plant disease.} Adapting these general-purpose models to agriculture has proven considerably harder, since agricultural imagery involves fine-grained morphological cues that general \gls{VLMs} are not trained to distinguish. This challenge has driven a progression of increasingly specialized adaptations, beginning with BioCLIP, which extends the CLIP framework to biological taxonomy and offers a foundation model for the tree of life that partially transfers to plant identification~\cite{BioCLIP}. Building on this taxonomic grounding, AgriCLIP narrows the focus further through domain-specialized cross-modal alignment tailored to agriculture and livestock applications~\cite{agriclip}, and SCOLD sharpens this specialization even more by refining vision-language alignment specifically for leaf disease identification~\cite{SCOLD}. Together, these efforts demonstrate measurable gains over general-purpose \gls{VLMs} as domain specificity increases, yet they remain anchored in classification-style supervision rather than open-ended diagnostic reasoning. This persistent reliance on classification underscores the need for benchmarks like TomaBench, which test deeper pathological understanding rather than surface-level recognition.

\section{Methodology} 
\label{methodology}

\subsection{TomaMMU Overview}
We built \gls{TomaMMU}, short for "Tomato leaf diseases MultiModal Understanding", a comprehensive dataset and a multimodal benchmark to evaluate the ability of \gls{VLMs} in understanding multiple tasks related to the phenotype and pathology of tomatoes. Two critical design choices were adopted for the data curation process: \textbf{Real-world distribution} images, questions, and answers are derived from real diagnoses and consulted by agricultural experts, and \textbf{Factual question} we employ \gls{OEQs}, providing the questions and requiring \gls{VLMs} to generate short-form responses that directly mention the key knowledge helpful to the user. In addition, we provided multiple-choice questions (MCQs) to align with most multimodal benchmarks. This combination released a comprehensive evaluation.

\noindent\textbf{Data Description} We designed a dataset-processing pipeline for human verification to convert biological and pathological information into high-quality question-answer (QA) pairs for agricultural knowledge evaluation. After filtering and processing, our \gls{TomaMMU} comprises: (1) a training dataset with 124,367 \gls{VQA} pairs specialized for fine-tuning \gls{VLMs} and (2) an evaluation set with 42,626 \gls{MCQs} and 37,816 \gls{OEQs} covering the knowledge types of tomato leaves called TomaBench. The large number of MCQs stems from our data construction strategy: each symptom in an image generates a separate MCQs with four options, while all symptoms in an image are merged into a single OEQs to produce a short answer.

\subsection{Dataset Curation}
To construct a robust multimodal question-answering model for tomato disease diagnosis, we developed the \gls{TomaMMU} dataset, comprising 28,808 with image size is $224\times224$ px images across 14 disease categories and one healthy category. To advance, we built TomaBench, a task-specific benchmark designed to evaluate the robustness of \gls{VLMs} in a zero-shot setting, which means the models are not fine-tuned on the QA pairs. As depicted in Fig. \ref{fig:data}, we outline a three-stage pipeline aspect follows: 

\begin{itemize}
    \item \textbf{Stage 1: Data Collection} The image data for this study consists of two parts: in-the-wild sources and in-house collections. These data were obtained from LeafNet \cite{LeafNet} and TOM2024 \cite{tom2024}, as shown in Fig. \ref{fig:Curation}b. Containing over 28,808 diseased tomato leaves representing 14 diseases and 25 tomato disease symptoms, \gls{TomaMMU} offers significantly greater scale and diversity than existing datasets. Previous studies relied predominantly on laboratory-acquired images characterized by uniform backgrounds. Although these models perform well on the internal dataset, in an uncontrolled environment, they often experience significant performance drops on out-of-distribution samples. To mitigate this generalization gap, \gls{TomaMMU} emphasizes in-situ data acquisition, with images captured directly from farms comprising the majority of the dataset. 
    \item \textbf{Stage 2: Human Annotation and Categorization} In Fig.  \ref{fig:data}b illustrates the diversity of diseases represented in \gls{TomaMMU}. From a pathological angle, the dataset encompasses a broad biological range; ``\textit{Yellow Leaf Curl Virus}'' and ``\textit{Septoria Leaf Spot}'' diseases dominate. This coarse-grained categorization is further refined through detailed taxonomic classification. To distinguish \gls{TomaMMU} from existing datasets, we implement an exhaustive annotation protocol. Instead of a simple static label, we extend the standard labels with structured metadata that includes detailed symptom descriptions, disease type, and quantity. To ensure the quality of the evaluation and precision of the ground-truth labels, mislabeled data is removed to ensure symptoms are visually identifiable. A human domain expert strictly executes the entire metadata extraction and annotation pipeline. Every data point undergoes a detailed manual verification process in which experts carefully filter and extract diagnostic knowledge, thereby ensuring the dataset's high accuracy and agronomic reliability. Through collecting and annotating, we harvest a dataset comprising 213,119 conversations. This distribution ensures that the data across the different categories is in general equilibrium. 
\end{itemize}

\begin{figure}[h]
    \centering
    \includegraphics[width=\linewidth]{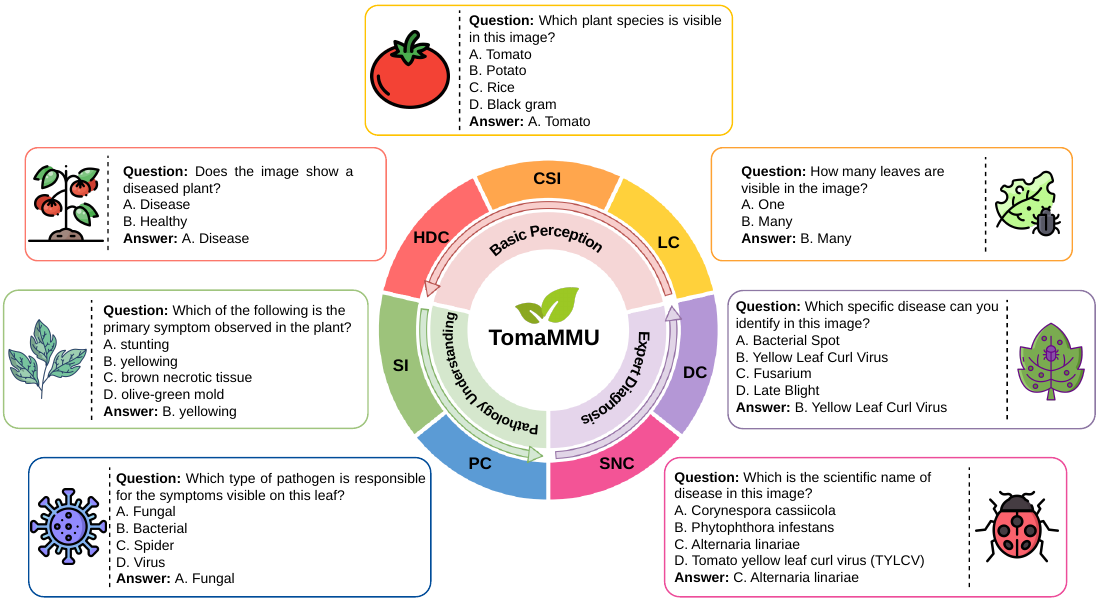}
    \caption{Visualization of the question tasks distribution in \gls{VQA} benchmark dataset. Each colored segment represents one of seven question types. The dataset includes HDC, SI, DC, LC, CSI, SNC and PC.}
    \label{fig:question_type_VQA}
\end{figure}

To assist \gls{VLMs} in handling complexity and a wide variety of real-world agricultural questions. We tag each question with an agriculture sub-domain. Guided by PlantVillageVQA's \cite{PlantVillageVQA} seasoned approach, we decided on the three categories "Basic Perception", "Pathology Understanding", and "Expert Diagnosis", as shown in the Fig. \ref{fig:data}c. The model will open with simple tasks, then move on to detailed level 2 analysis and verification, focusing on detecting symptoms and pathogens, and finally to higher-level inference and judgment, handling more complex questions. This part requires the model to synthesize knowledge and reason to answer. This design mitigates the hallucination problem typical of \gls{VLMs}. Then, to systematically extract knowledge, we organize agricultural knowledge into categories of Healthy-Diseased Classification (HDC), Leaf Classification (LC), Disease Classification (DC), Crop Species Identification (CSI), Scientific Name Classification (SNC), Pathogen Classification (PC) and Symptom Identification (SI), as shown in Fig. \ref{fig:question_type_VQA}. This structured design evaluates model performance by grouping these seven tasks into the three aforementioned hierarchical sub-domains:
\begin{itemize}
    \item \textbf{Basic perception:} Provides basic classification of the presence of pathology (HDC), leaf condition (LC), and evaluates host crop species identification (CSI).
    \item \textbf{Pathology Understanding:} Focuses on fine-grained manifestations such as spots, chlorosis, and mosaic symptoms (SI), which help identify the pathogen category (PC).
    \item \textbf{Expert Diagnosis:} Target specific pathological condition (DC) and predict the scientific name of the disease on the leaf (SNC).
\end{itemize}

\noindent\textbf{Stage 3: QA Generation} For rigorous evaluation, \gls{TomaMMU} adopts \gls{MCQs} and \gls{OEQs}. Using the dataset, a small, balanced subset, TomaBench, is selected for evaluation, while the remaining data is used for fine-tuning and training. All questions are designed to be visual-dependent and ensure that the selected conversations closely resemble real-life conversations between farmers and experts. To balance across different agricultural domains and knowledge types, we prioritize an even distribution of question types, as shown in Fig. \ref{fig:Curation}c. To generate \gls{MCQs}, depending on the knowledge type of the fact, we generate three wrong answers and form one correct answer, using the extracted fact as the ground truth for the model to predict. With \gls{OEQs}, the models are set up to offer short answers and uses Gemini 2.0 Flash \cite{Gemini2.0} as a judge to validate generated answers and ground truth labels to specify semantic accuracy and correctness of VLMs.

\subsection{Additional Properties of \gls{TomaMMU}}
\begin{figure}[t]
    \centering
    \includegraphics[
    width=\linewidth]{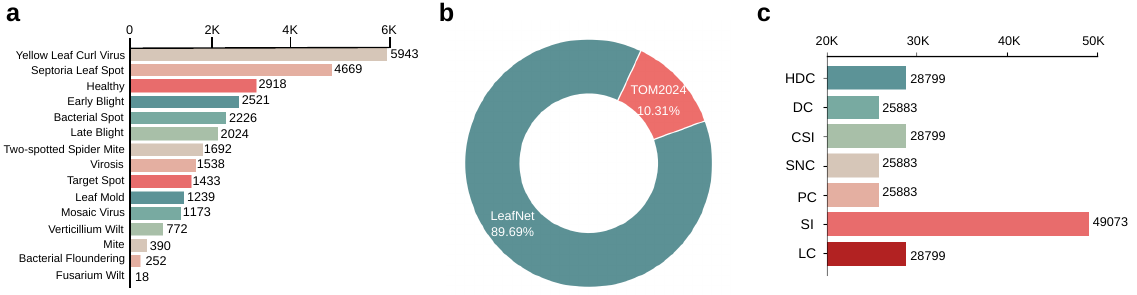}
    \caption{Visualization of key components of \gls{TomaMMU}. \textbf{a,} Distribution of number classes of disease in the \gls{TomaMMU}. \textbf{b,} \gls{TomaMMU} data sources. \textbf{c,} Number of \gls{VQA} on different question types.}
    \label{fig:Curation}
\end{figure}

\textbf{Distribution and Coverage} In Fig. \ref{fig:Curation}, we show the distribution of the number of classes and knowledge types in \gls{TomaMMU}. We have two key observations: (1) In the raw image distribution, natural factual events are heavily skewed toward a few major diseases, leaving other important conditions underrepresented, resulting in a severe imbalance across disease categories. To address this, we developed a structured \gls{VQA} framework in which a single image is associated with up to seven distinct question types, ranging from basic perception to expert-level diagnosis. This strategy significantly enriches the density of supervision signals and balances the representation of agricultural knowledge. As demonstrated in Fig. \ref{fig:Curation}c, these generated \gls{VQA} conversations exhibit a markedly more uniform distribution across task categories, facilitating a more equitable and comprehensive assessment of \gls{VLMs} capabilities. This is particularly important for our evaluation dataset, where we aim to propose a comprehensive standard that tests different aspects of the model without inductive bias. 

\noindent \textbf{Realistic Images} Our \gls{TomaMMU} is also different from photography-level images, laboratory datasets like PlantVillage, or textbook or web-document such as manually curated CROP \cite{crop} and \gls{CDDM} \cite{cddm}. \gls{TomaMMU} is a conjunction of a large portion of data collected directly in the field (farm) with some data from a controlled environment (laboratory). Our benchmark also emphasizes training \gls{VLMs} to align complex morphological patterns (such as sores, spots, colors) with in-depth textual descriptions.

\section{Experimental Result}
We conducted comprehensive evaluations to demonstrate the efficacy of domain-specific fine tuning in agricultural tasks and to establish a robust comparative baseline in a variety of different \gls{VLMs}. Primarily, we present refined results on TomaLLaVA, a multimodal generative framework that jointly leverages visual and textual information to perform accurate and interpretable tomato disease diagnosis. Furthermore, we benchmarked several state-of-the-art \gls{VLMs}. Our evaluations focused primarily on zero-shot performance, using either publicly available APIs or author-provided checkpoints to reflect the inherent capabilities of each model without task-specific training. All experiments were conducted using NVIDIA RTX 4060 Ti GPUs, and more detailed information on model configuration, API choices, and implementation details is included in the supplementary documentation.


\noindent\textbf{SOTA competing models.} We assess a broad range of state-of-the-art \gls{VLMs} to ensure robust comparisons. For open-sourced models, we prioritize architectures with comparable parameter numbers for a fair and meaningful comparison. We identify three categories of models: Proprietary \gls{VLMs}, generative FMs, and CLIP-based models. \textit{Proprietary VLMs} is Gemini 2.0 Flash \cite{Gemini2.0}, \textit{General FMs}. We include eight \gls{VLMs} (InternVL3 \cite{InternVL3}, Qwen2.5-3B \cite{Qwen2.5}, Qwen3-2B \cite{Qwen3}, Gemma4-E4B \cite{gemma4-wang}, SmolVLM2-2.2B \cite{smol}, LLaVA1.5 \cite{LlaVA} and LFM2-VL \cite{LFM2}). \textit{CLIP-based models.} we utilize (CLIP \cite{clip}, SigLIP2 \cite{SigLIP2}), and two \textit{domain-specific \gls{VLMs}} (BioCLIP \cite{BioCLIP} and SCOLD \cite{SCOLD}), which are models refined from existing foundational models to optimize for specific tasks. All were evaluated without task-specific fine-tuning to assess out-of-the-box generalization.

\subsection{Evaluation Metrics}

To ensure robustness and fairness, we evaluate the models on \gls{MCQs} and \gls{OEQs}. 

\noindent \textbf{For \gls{MCQs}}, we report Accuracy (Acc) and F1 score (F1) as the primary evaluation metric. We score the model's response by matching its predicted option to the ground truth. 

\noindent \textbf{For \gls{OEQs}}, we first utilize ROUGE-L \cite{rouge} to comprehensively assess the lexical and structural similarity between generated responses and human-annotated data. Furthermore, we implement the LLM-as-judge methodology using the GPT score to grade answers semantically. The short-form responses correspond to the questions, which normally contain several words, and are evaluated directly against the correct answers. Specifically, our \gls{LLM} judge follows a three-step process: (a) dividing both predicted and ground-truth responses into individual statements; (b) grading each of them as a short-form response; and (c) normalizing the grades according to the number of statements per question. This allows us to calculate the final harmonic mean of the semantic scores (1-5), which is designed to punish complete errors (score 1) severely. Detailed construction and specifications are provided in Section 4.4.


\begin{table*}[htbp] 

\centering

\resizebox{\textwidth}{!}{

\begin{tabular}{l c c c c c c c c c c c c c c c c}

\toprule

\multirow{2}{*}{\textbf{Model}} & \multicolumn{2}{c}{\textbf{HDC}} & \multicolumn{2}{c}{\textbf{DC}} & \multicolumn{2}{c}{\textbf{CSI}} & \multicolumn{2}{c}{\textbf{SNC}} & \multicolumn{2}{c}{\textbf{PC}} & \multicolumn{2}{c}{\textbf{SI}} & \multicolumn{2}{c}{\textbf{LC}} & \multicolumn{2}{c}{\textbf{Average}} \\
\cmidrule(lr){2-3} \cmidrule(lr){4-5} \cmidrule(lr){6-7} \cmidrule(lr){8-9} \cmidrule(lr){10-11} \cmidrule(lr){12-13} \cmidrule(lr){14-15} \cmidrule(lr){16-17}
 & Acc & F1 & Acc & F1 & Acc & F1 & Acc & F1 & Acc & F1 & Acc & F1 & Acc & F1 & Acc & F1 \\ 
\midrule
\multicolumn{17}{c}{\textit{Proprietary \gls{VLMs}}} \\ \midrule
Gemini 2.0 Flash \cite{Gemini2.0} & 91.0 & 62.6 & 47.5 & 36.2 & 92.9 & 92.9 & 51.6 & 51.6 & 70.5 & 70.5 & 44.0 & 45.4 & 94.6 & 86.9 & 70.3 & 63.7 \\

\midrule

\multicolumn{17}{c}{\textit{Generative \gls{VLMs}}} \\ \midrule

SmolVLM2-2.2B \cite{smol} & 36.3 & 23.0 & 33.3 & 31.3 & 36.3 & 33.4 & 42.5 & 42.6 & 40.1 & 37.7 & 36.4 & 35.5 & 92.9 & 56.3 & 45.4 & 37.1 \\ 
Gemma4-E4B \cite{gemma4-wang} & 11.6 & 11.0 & 30.9 & 19.8 & 27.4 & 18.8 & 24.4 & 14.3 & 24.5 & 20.9 & 25.3 & 19.1 & 29.5 & 27.2 & 24.8 & 18.7 \\ 
InternVL3-1B \cite{InternVL3} & 89.9 & 47.3 & 28.6 & 28.5 & 13.7 & 13.6 & 30.7 & 30.5 & 50.8 & 50.6 & 37.5 & 37.6 & 93.7 & 84.7 & 49.3 & 41.8 \\ 
InternVL3-2B \cite{InternVL3} & 82.1 & 43.6 & 34.2 & 34.1 & 46.2 & 46.1 & 42.7 & 42.6 & 51.4 & 51.5 & 32.4 & 32.4 & 94.4 & 85.3 & 54.8 & 47.9 \\ 
Qwen2.5-3B \cite{Qwen2.5} & 83.8 & 64.1 & 40.6 & 39.8 & 62.9 & 62.9 & 47.5 & 47.3 & 51.0 & 51.0 & 32.5 & 32.3 & 92.6 & 52.6 & 58.7 & 50.0 \\ 
Qwen3-2B \cite{Qwen3} & 91.6 & 48.0 & 45.6 & 44.2 & 60.9 & 61.0 & 32.6 & 32.6 & 55.9 & 55.2 & 33.9 & 32.3 & 93.7 & 84.2 & 59.2 & 51.1 \\ 

LLaVA-1.5-7B \cite{LlaVA} & 89.9 & 47.3 & 34.8 & 23.6 & 50.2 & 49.1 & 24.8 & 10.3 & 45.2 & 42.5 & 26.7 & 16.4 & 93.2 & 80.1 & 52.1 & 38.5 \\ 

LFM2.5-VL-1.6B \cite{LFM2} & 90.9 & 75.3 & 41.9 & 41.6 & 76.5 & 76.4 & 38.1 & 35.3 & 54.1 & 53.8 & 33.0 & 29.7 & 94.3 & 42.8 & 61.3 & 50.7 \\ 

\midrule

\multicolumn{17}{c}{\textit{CLIP-based Models}} \\ \midrule

SigLIP2 \cite{SigLIP2} & 76.1 & 63.7 & 25.6 & 25.2 & 30.1 & 30.1 & 27.8 & 27.8 & 11.2 & 11.1 & 28.6 & 28.6 & 93.1 & 81.1 & 41.8 & 38.2 \\ 

CLIP-ViT-L-14 \cite{clip} & 90.2 & 78.1 & 36.9 & 36.5 & 54.2 & 54.2 & 29.7 & 29.7 & 51.3 & 51.2 & 26.1 & 26.1 & 59.7 & 52.4 & 49.7 & 46.9 \\ 

SCOLD \cite{SCOLD} & 55.1 & 48.7 & 74.6 & 74.2 & \textbf{98.0} & \textbf{98.0} & 49.3 & 49.3 & 47.4 & 47.4 & 52.1 & 52.1 & 44.9 & 39.9 & 60.2 & 58.5 \\

BioCLIP-ViT-B-32 \cite{BioCLIP} & 90.0 & 51.1 & 44.5 & 26.5 & 56.4 & 2.7 & 43.4 & 27.0 & 40.8 & 20.0 & 23.2 & 15.4 & 93.2 & 80.4 & 55.9 & 31.9 \\ 

\midrule

{\textbf{TomaLLaVA }} & {\textbf{98.9}} & {\textbf{98.9}} & {\textbf{96.4}} & {\textbf{96.4}} & - & - & {\textbf{95.5}} & {\textbf{95.5}} & {\textbf{98.9}} & {\textbf{98.9}} & {\textbf{91.2}} & {\textbf{88.7}} & {\textbf{99.2}} & {\textbf{99.2}} & {\textbf{96.1}} & {\textbf{95.8}} \\ 

\bottomrule

\end{tabular}%

}

\caption{Performance of \gls{VLMs} on our TomaBench \gls{MCQs}. Our evaluation set poses great challenges to existing large \gls{VLMs}. TomaLLaVA ignores the CSI task for non-image crops in TomaBench.}

\label{tab:model_comparison}

\end{table*}


\subsection{Zero-shot Evaluation Results}
We present a comprehensive zero-shot comparison of various \gls{VLMs} in Table \ref{tab:model_comparison}. Evaluated specifically on the TomaBench for Generative FMs, CLIP-based models, Fine-tuned models, and Domain-specific \gls{VLMs}. 

\noindent \textbf{This presents significant challenges for existing models}
Beyond simple classification, diagnostic support systems must possess reasoning capabilities. TomaBench proved to be a very challenging benchmark for all evaluated models, and even the most advanced systems achieved average performance levels as shown in Table \ref{tab:OEQ}. These challenges are noteworthy, particularly for \gls{OEQs}, where models must accurately recall knowledge without fine-tuning on question-answer pairs. These observations also highlight the need to improve the agricultural expertise of \gls{VLMs} and to incorporate agricultural data into \gls{VLMs} training, where our \gls{TomaMMU} could be useful.

\noindent \textbf{Performance Across Question Types} On \gls{MCQs}, We assessed model performance on closed-form QA multiple-choice questions, which are significantly easier than \gls{OEQs}. This protocol tests the fit between visual features and complex agricultural linguistic concepts, measuring the model's ability to perform expert-level inference without task-specific training. The level of accuracy is also higher than the agricultural standards set by the \gls{OEQs} in the responses. However, there is significant variation among models in which tasks are the most difficult. In particular, PC, SI, and SNC tasks struggle with agricultural diagnostics. This reflects the inherent difficulty of these tasks, which requires the model to process detailed morphological descriptions. In contrast, binary HDC and LC maintain a concise and uniform prompt structure, consistent with their high-level screening objective. Suggests that specialized agricultural knowledge and visual understanding capabilities are not uniformly distributed across model architectures.

\begin{table}[htbp]
\centering
\footnotesize
\setlength{\tabcolsep}{4pt} 
\renewcommand{\arraystretch}{1.2} 
\resizebox{\textwidth}{!}{
\begin{tabular}{l | c c c c c c c | c | c}
\hline
& \multicolumn{9}{c}{\textbf{TomaBench-OEQs}} \\ \cline{2-10}

& \rotatebox{90}{HDC} \rule[-8pt]{0pt}{46pt}
& \rotatebox{90}{DC} 
& \rotatebox{90}{CSI} 
& \rotatebox{90}{SNC} 
& \rotatebox{90}{PC} 
& \rotatebox{90}{SI} 
& \rotatebox{90}{LC} 
& \rotatebox{90}{Avg.} 
& \rotatebox{90}{GPT} \\ \hline

SmolVLM2-2.2B \cite{smol}  & 0.458 & 0.146 & 0.043 & 0.035 & 0.348 & 0.050 & 0.911 & 0.284 & $\sim$2.3 \\
Gemma4-E4B \cite{gemma4-wang}   & 0.694 & 0.191 & 0.017 & 0.005 & 0.323 & 0.073 & 0.707 & 0.287 & $\sim$2.3 \\
InternVL3 1B \cite{InternVL3} & 0.816 & 0.016 & 0.001 & 0.001 & 0.381 & 0.142 & 0.840 & 0.314 & $\sim$2.1 \\
InternVL3 2B \cite{InternVL3} & 0.874 & 0.255 & 0.055 & 0.060 & 0.357 & 0.065 & 0.905 & 0.367 & $\sim$2.2 \\
Qwen2.5 3B \cite{Qwen2.5}  & 0.790 & 0.234 & 0.034 & 0.048 & 0.267 & 0.202 & 0.796 & 0.339 & $\sim$2.4 \\
Qwen3 2B \cite{Qwen3}    & 0.899 & 0.182 & 0.341 & 0.099 & 0.071 & 0.203 & 0.869 & 0.380 & $\sim$2.7 \\
LLaVA-OV \cite{LlaVA}    & 0.914 & 0.023 & 0.111 & 0.000 & 0.005 & 0.071 & 0.906 & 0.290 & $\sim$2.6 \\
LFM-VL 2.5 \cite{LFM2}  & 0.929 & 0.141 & 0.715 & 0.019 & 0.319 & 0.166 & 0.895 & 0.455 & $\sim$2.6 \\
\hline 
\end{tabular}
}
\caption{Performance of \gls{VLMs} on our TomaBench \gls{OEQs}. Our evaluation set poses great challenges to existing large \gls{VLMs}.}
\label{tab:OEQ}
\end{table}

\subsection{Fine-tuning Result}
\begin{figure}[t]
    \centering
    \includegraphics[width=\linewidth]{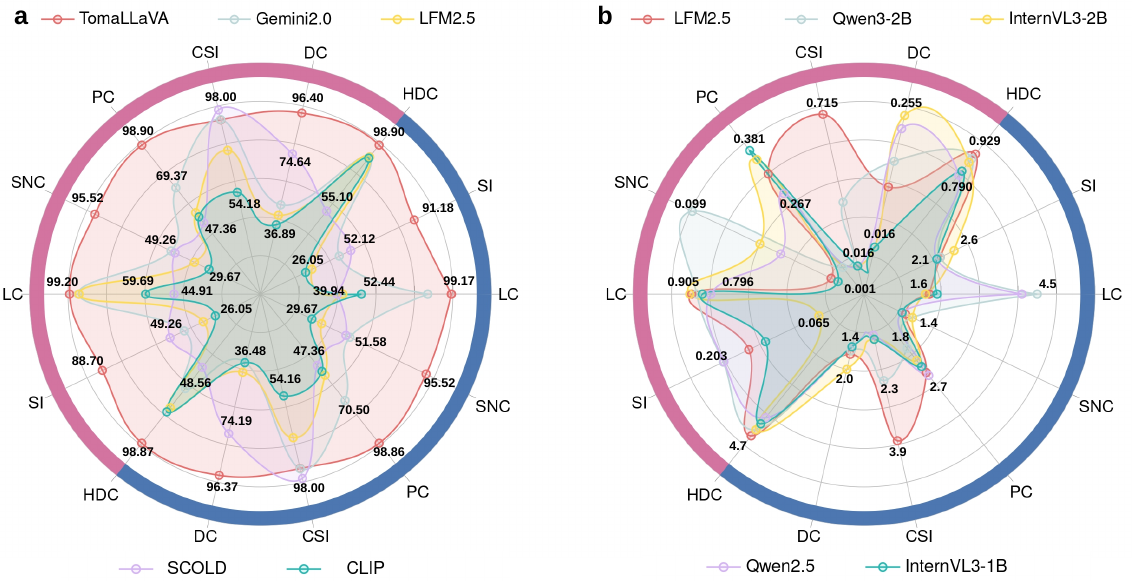}
    \caption{Summary of model results on MCQs and OEQs. \textbf{a,} Performance of various foundation models across seven tomato disease understanding tasks (\textcolor[HTML]{D373A0}{\textbf{pink}} indicates Acc, and \textcolor[HTML]{4D77B0}{\textbf{blue}} denotes F1). Performance of Top-5 SOTA models on \gls{MCQs}, showing that TomaLLaVA maintains the highest overall performance. b, Comparison of the Top-5 state-of-the-art models on \gls{OEQs} (\textcolor[HTML]{D373A0}{\textbf{pink}} indicates the average of ROUGE-L, and \textcolor[HTML]{4D77B0}{\textbf{blue}} denotes the GPT score.}
    \label{fig:AccF1}
\end{figure}

Although agricultural \gls{VLMs} have made progress, they still fail to exploit the specific characteristics of individual diseases or species. As a result, these models often lack the detailed domain knowledge required to distinguish apparently similar tomato diseases and to support reasoning-based diagnosis across diverse real-world agricultural conditions, and tomato leaves disease data is the initial foundation. Notably, even highly advanced foundation models such as Gemini 2.0 Flash \cite{Gemini2.0} only achieved 70.30\% on the overall dataset; therefore, the need for a model fine-tuned specifically on this dataset was paramount. This performance gap illustrates the requirement to train a domain model for \gls{TomaMMU}. This comparative evaluation demonstrates TomaLLaVA's superior ability to generate accurate, contextually rich disease diagnoses and symptom explanations beyond what standard classification-based approaches can provide.
\noindent In Fig. \ref{fig:AccF1} summarizes the performance of all models and top models across seven question types. Overall, TomaLLaVA achieves the highest performance across all tasks, with an average Acc of 96.09\% and an F1 of 0.958, substantially outperforming all competing models. As shown in Table \ref{tab:model_comparison}. Compared with the best \gls{VLMs} (LFM2.5, 61.26\% Acc, 0.507 F1), TomaLLaVA gains +34.8\% Acc and +0.451 F1, even in Fine-tuned \gls{VLMs} (AgriCLIP, 93.73\% Acc, 0.948 F1), TomaLLaVA gains +2.4\% Acc and +0.01 F1. Overall, the excellent TomaLLaVA results cover all models and deliver near-perfect results for perception tasks and strong fine-grained disease reasoning, confirming its ability to emulate expert-level diagnostic reasoning through multimodal integration.

\subsection{Error Analysis for \gls{OEQs}}
To achieve an in-depth understanding of the internal mechanisms and limitations of \gls{VLMs} when answering \gls{OEQs}, we conducted automated error analysis and qualitative evaluation. Unlike some previous studies that relied entirely on manual scoring, we applied an LLM-as-a-judge method, using the GPT score to evaluate and score model predictions based on the reference answer (Ground Truth) on a scale of 1 to 5. Our prompt for Gemini 2.0 Flash is presented in Table \ref{tab:prompt_template}. 

Table \ref{tab:OEQ} showed that all models encountered limitations and achieved only scores $\approx2.5$. Therefore, we classified and clarified the nature of the errors corresponding to these three score scenarios.

\noindent \textbf{Perceptual Error \& Refusal to Answer} (Score 1) This delegation for dead failure in the process question type: the \gls{VLMs} lacks knowledge or visual grounding, giving completely wrong, irrelevant, or hallucinatory answers. For example, mistaking ``\textit{Tomato}'' for ``\textit{Eucalyptus}'' or ``\textit{Virus}'' for ``\textit{Bacteria}''. Moreover, instead of an analytical image, Gemma4-E4B \cite{gemma4-wang} repeated allegations, stereotypes, and negative responses such as ``\textit{No plant is visible}'' for SI or ``\textit{No visible disease}'' for DC, even though the image clearly shows signs of the disease. The system failure rendered the predictions completely worthless. \

\noindent\textbf{Core Miss \& Major Inaccuracy} (Score 2) The answer is partially relevant but severely inaccurate or misses the core idea. In these cases, the model may identify relevant visual features and strengthen them, but still fail to reach the correct answer (e.g., describing ``\textit{Brown Spot}'' instead of ``\textit{Early Leaf Blight}''). In other cases, the model only identified secondary symptoms ``\textit{wilting}'' while overlooking the dangerous primary symptoms ``\textit{death}'' and ``\textit{stunting}''. 

\noindent\textbf{Granularity \& Detail Omission} (Score 3) The answer is generally correct but lacks specific details or contains minor factual errors. This category highlights limitations in deep agricultural knowledge. For instance, recognizing discoloration ``\textit{yellowing}'' but overlooking accompanying structural deformation ``\textit{curling}'', or correctly identifying the type of damage ``\textit{leaf spots}'' but lacking detail about the color ``\textit{reddish-brown}''.


\subsection{Training result}
\textbf{Experiment Setting} 
We conducted training TomaLLaVA, which uses SCOLD \cite{SCOLD} as the image encoder and LFM2.5-1.2B-Instruct \cite{LFM2} as the base \gls{LLM}. Our training lasts 20 epochs using Low-Rank Adaptation (LoRA) \cite{lora}, a learning rate of 1e-4, a weight decay of 0.01, and a batch size of 32. The fine-tuning process takes on NVIDIA RTX 4060 Ti GPUs.

\setlength{\intextsep}{2pt}
\begin{wrapfigure}[16]{r}{0.5\columnwidth}
    \includegraphics[width=\linewidth]{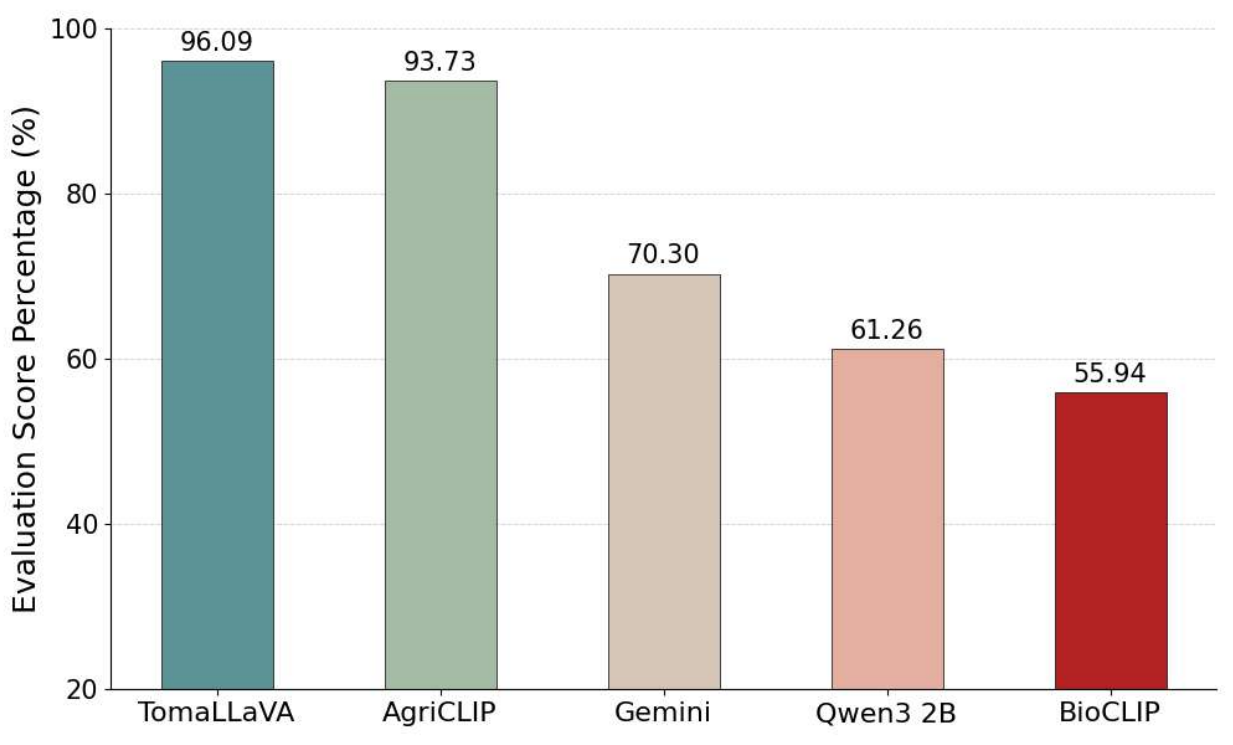}
    \caption{Average evaluation scores on \gls{MCQs}. Fine-tuning TomaLLaVA significantly enhances its understanding of agricultural knowledge, demonstrating the effectiveness of our development set compared with four state-of-the-art models.}
    \label{fig:Fine_tuned}
\end{wrapfigure}

\textbf{Fine-tuning Performance.} Fig. \ref{fig:Fine_tuned} shows that the fine-tuning with our knowledge base generates a significant performance boost, improving the capability of understanding the images and correctly responding with agriculture knowledge, leading to an average improvement of up to 2.36\% on \gls{MCQs}. This performance boost demonstrates the effectiveness of our large-scale knowledge base in improving \gls{VLMs} and the demand to collect agricultural-related data for future \gls{VLMs}. However, for questions on pathology understanding and expert diagnosis, the level of improvement is smaller, indicating the complexity of multi-statement questions and underscoring the need for more data collection and better model training beyond simple fine-tuning. 

\begin{figure}[h]
    \centering
    \includegraphics[width=\linewidth]{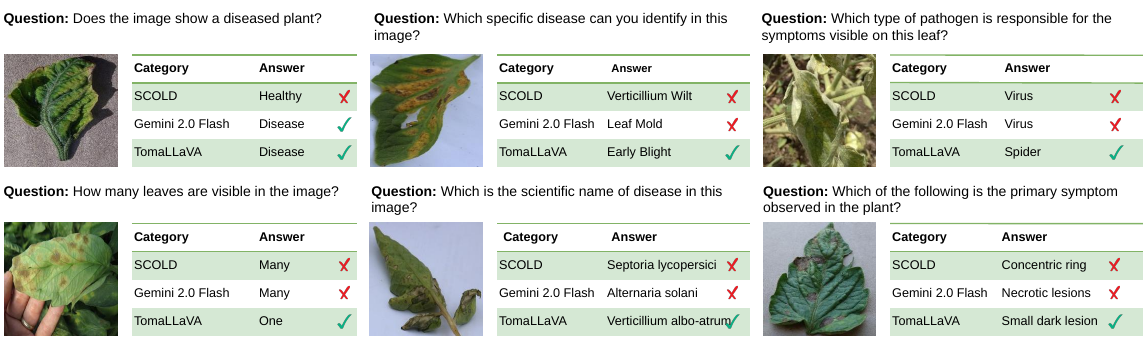}
    \caption{The effectiveness of TomaLLaVA fine-tuning on OEQ examples. After simple fine-tuning, TomaLLaVA can accurately identify issues that SCOLD~\cite{SCOLD} and Gemini 2.0 Flash \cite{Gemini2.0} failed to recognize in zero-shot scenarios.}
    \label{fig:Analysic}
\end{figure}

\textbf{Failure Exemplars.} Fig. \ref{fig:Analysic} indicated that the most persistent bottlenecks lie in agronomic reasoning, which requires interpretation of biological and agronomic context, such as symptoms, pathogens, and disease progression. These limitation stem from general datasets lacking specialized knowledge, diversity, and realism, making the models susceptible to illusions and mispredictions. For instance, when asked to identify a specific disease from an image of a leaf with characteristic necrotic spots, generals VLMs lacked of agronomic refinement gave incorrect diagnoses such as SCOLD \cite{SCOLD} predicted ``\textit{Verticillium Wilt}'' and Gemini 2.0 Flash \cite{Gemini2.0} predicted ``\textit{Leaf Mold}''. In contrast, TomaLLaVA accurately applied specialized knowledge to correctly identify ``\textit{Early Blight}''. Additionally, for simple tasks like LC, foundation models such as SCOLD and Genini 2.0 Flash still focus more on the background than on the main object, as their predictions are ''Many".

\section{Conclusion}
In this study, we introduce \gls{TomaMMU}, a large-scale dataset and benchmark for tomato disease \gls{VLMs} in agriculture, a field that requires high levels of expert knowledge and accurate interpretation of visuals. Our dataset comprises over 213,119 human-annotated image metadata pairs. Applying a three-stage curation framework, we construct a high-quality \gls{TomaMMU} evaluation set and demonstrate its crucial role in improving performance of multimodal models. To support model development, we propose TomaBench set, including 42,626 question-answer pairs to evaluate multimodal ability. Our contributions encompass not only the dataset but also a structured fine-tuning strategy and a human-annotated gold standard split to foster future research in agricultural technology and improve the capabilities of \gls{AI} systems with complex reasoning requirements for tomato leaves disease, as well as a platform to build and deploy to other datasets in agriculture. 
 
\section*{Acknowledgment}
Khang Nguyen Quoc was supported by the Hyundai Motor Chung Mong-Koo Foundation Global Scholarship (GSS-25-02120).


%
%
\bibliographystyle{splncs04}
\bibliography{main}

@String(ECCV  = {Eur. Conf. Comput. Vis.})

@String(ECCV  = {ECCV})

@article{George2025,
  title = {Past,  present and future of deep plant leaf disease recognition: A survey},
  volume = {234},
  ISSN = {0168-1699},
  url = {http://dx.doi.org/10.1016/j.compag.2025.110128},
  DOI = {10.1016/j.compag.2025.110128},
  journal = {Computers and Electronics in Agriculture},
  publisher = {Elsevier BV},
  author = {George,  Romiyal and Thuseethan,  Selvarajah and Ragel,  Roshan G. and Mahendrakumaran,  Kayathiri and Nimishan,  Sivaraj and Wimalasooriya,  Chathrie and Alazab,  Mamoun},
  year = {2025},
  month = July,
  pages = {110128}
}

@misc{tom2024,
  doi = {10.17632/3D4YG89RTR.1},
  url = {https://data.mendeley.com/datasets/3d4yg89rtr/1},
  author = {Appiah,  Obed and Hackman,  Kwame Oppong  and Diallo,  Belko Abdoul Aziz and Ogunjobi,  Kehinde O and Ouedraogo,  Valentin and Bebe,  Momo and SON,  Diakalia},
  title = {TOM2024},
  publisher = {Mendeley Data},
  year = {2024}
}

@inproceedings{PlantDoc,
  series = {CoDS COMAD 2020},
  title = {PlantDoc: A Dataset for Visual Plant Disease Detection},
  url = {http://dx.doi.org/10.1145/3371158.3371196},
  DOI = {10.1145/3371158.3371196},
  booktitle = {Proceedings of the 7th ACM IKDD CoDS and 25th COMAD},
  publisher = {ACM},
  author = {Singh,  Davinder and Jain,  Naman and Jain,  Pranjali and Kayal,  Pratik and Kumawat,  Sudhakar and Batra,  Nipun},
  year = {2020},
  month = Jan,
  pages = {249–253},
  collection = {CoDS COMAD 2020}
}

@article{FieldPlant,
  title = {FieldPlant: A Dataset of Field Plant Images for Plant Disease Detection and Classification With Deep Learning},
  volume = {11},
  ISSN = {2169-3536},
  url = {http://dx.doi.org/10.1109/ACCESS.2023.3263042},
  DOI = {10.1109/access.2023.3263042},
  journal = {IEEE Access},
  publisher = {Institute of Electrical and Electronics Engineers (IEEE)},
  author = {Moupojou,  Emmanuel and Tagne,  Appolinaire and Retraint,  Florent and Tadonkemwa,  Anicet and Wilfried,  Dongmo and Tapamo,  Hyppolite and Nkenlifack,  Marcellin},
  year = {2023},
  pages = {35398–35410}
}

@article{PlantVillage,
  title = {Using Deep Learning for Image-Based Plant Disease Detection},
  volume = {7},
  ISSN = {1664-462X},
  url = {http://dx.doi.org/10.3389/fpls.2016.01419},
  DOI = {10.3389/fpls.2016.01419},
  journal = {Frontiers in Plant Science},
  publisher = {Frontiers Media SA},
  author = {Mohanty,  Sharada P. and Hughes,  David P. and Salathé,  Marcel},
  year = {2016},
  month = Sept 
}

@misc{VQAv2,
  doi = {10.48550/ARXIV.1612.00837},
  url = {https://arxiv.org/abs/1612.00837},
  author = {Goyal,  Yash and Khot,  Tejas and Summers-Stay,  Douglas and Batra,  Dhruv and Parikh,  Devi},
  title = {Making the V in VQA Matter: Elevating the Role of Image Understanding in Visual Question Answering},
  publisher = {arXiv},
  year = {2016},
  copyright = {arXiv.org perpetual,  non-exclusive license}
}

@misc{GQA,
  doi = {10.48550/ARXIV.1902.09506},
  url = {https://arxiv.org/abs/1902.09506},
  author = {Hudson,  Drew A. and Manning,  Christopher D.},
  title = {GQA: A New Dataset for Real-World Visual Reasoning and Compositional Question Answering},
  publisher = {arXiv},
  year = {2019},
  copyright = {Creative Commons Attribution 4.0 International}
}

@article{ScienceQA,
  title = {ScienceQA: a novel resource for question answering on scholarly articles},
  volume = {23},
  ISSN = {1432-1300},
  url = {http://dx.doi.org/10.1007/s00799-022-00329-y},
  DOI = {10.1007/s00799-022-00329-y},
  number = {3},
  journal = {International Journal on Digital Libraries},
  publisher = {Springer Science and Business Media LLC},
  author = {Saikh,  Tanik and Ghosal,  Tirthankar and Mittal,  Amish and Ekbal,  Asif and Bhattacharyya,  Pushpak},
  year = {2022},
  month = July,
  pages = {289–301}
}

@misc{tong2024eyes,
  doi = {10.48550/ARXIV.2401.06209},
  url = {https://arxiv.org/abs/2401.06209},
  author = {Tong,  Shengbang and Liu,  Zhuang and Zhai,  Yuexiang and Ma,  Yi and LeCun,  Yann and Xie,  Saining},
  title = {Eyes Wide Shut? Exploring the Visual Shortcomings of Multimodal LLMs},
  publisher = {arXiv},
  year = {2024},
  copyright = {Creative Commons Attribution 4.0 International}
}

@misc{yue2024mmmu,
  doi = {10.48550/ARXIV.2311.16502},
  url = {https://arxiv.org/abs/2311.16502},
  author = {Yue,  Xiang and Ni,  Yuansheng and Zhang,  Kai and Zheng,  Tianyu and Liu,  Ruoqi and Zhang,  Ge and Stevens,  Samuel and Jiang,  Dongfu and Ren,  Weiming and Sun,  Yuxuan and Wei,  Cong and Yu,  Botao and Yuan,  Ruibin and Sun,  Renliang and Yin,  Ming and Zheng,  Boyuan and Yang,  Zhenzhu and Liu,  Yibo and Huang,  Wenhao and Sun,  Huan and Su,  Yu and Chen,  Wenhu},
  title = {MMMU: A Massive Multi-discipline Multimodal Understanding and Reasoning Benchmark for Expert AGI},
  publisher = {arXiv},
  year = {2023},
  copyright = {arXiv.org perpetual,  non-exclusive license}
}

@inbook{cddm,
  title = {A Multimodal Benchmark Dataset and Model for Crop Disease Diagnosis},
  ISBN = {9783031730160},
  ISSN = {1611-3349},
  url = {http://dx.doi.org/10.1007/978-3-031-73016-0_10},
  DOI = {10.1007/978-3-031-73016-0_10},
  booktitle = {Computer Vision – ECCV 2024},
  publisher = {Springer Nature Switzerland},
  author = {Liu,  Xiang and Liu,  Zhaoxiang and Hu,  Huan and Chen,  Zezhou and Wang,  Kohou and Wang,  Kai and Lian,  Shiguo},
  year = {2024},
  month = Oct,
  pages = {157–170}
}

@inbook{AgriBench,
  title = {AgriBench: A Hierarchical Agriculture Benchmark for Multimodal Large Language Models},
  ISBN = {9783031918353},
  ISSN = {1611-3349},
  url = {http://dx.doi.org/10.1007/978-3-031-91835-3_14},
  DOI = {10.1007/978-3-031-91835-3_14},
  booktitle = {Computer Vision – ECCV 2024 Workshops},
  publisher = {Springer Nature Switzerland},
  author = {Zhou,  Yutong and Ryo,  Masahiro},
  year = {2025},
  pages = {207–223}
}

@misc{LeafNet,
  doi = {10.48550/ARXIV.2602.13662},
  url = {https://arxiv.org/abs/2602.13662},
  author = {Quoc,  Khang Nguyen and Dao,  Phuong D. and Quach,  Luyl-Da},
  title = {LeafNet: A Large-Scale Dataset and Comprehensive Benchmark for Foundational Vision-Language Understanding of Plant Diseases},
  publisher = {arXiv},
  year = {2026},
  copyright = {Creative Commons Attribution Non Commercial No Derivatives 4.0 International}
}

@misc{lora,
  doi = {10.48550/ARXIV.2106.09685},
  url = {https://arxiv.org/abs/2106.09685},
  author = {Hu,  Edward J. and Shen,  Yelong and Wallis,  Phillip and Allen-Zhu,  Zeyuan and Li,  Yuanzhi and Wang,  Shean and Wang,  Lu and Chen,  Weizhu},
  title = {LoRA: Low-Rank Adaptation of Large Language Models},
  publisher = {arXiv},
  year = {2021},
  copyright = {arXiv.org perpetual,  non-exclusive license}
}

@misc{smol,
  doi = {10.48550/ARXIV.2502.02737},
  url = {https://arxiv.org/abs/2502.02737},
  author = {Allal,  Loubna Ben and Lozhkov,  Anton and Bakouch,  Elie and Blázquez,  Gabriel Martín and Penedo,  Guilherme and Tunstall,  Lewis and Marafioti,  Andrés and Kydlíček,  Hynek and Lajarín,  Agustín Piqueres and Srivastav,  Vaibhav and Lochner,  Joshua and Fahlgren,  Caleb and Nguyen,  Xuan-Son and Fourrier,  Clémentine and Burtenshaw,  Ben and Larcher,  Hugo and Zhao,  Haojun and Zakka,  Cyril and Morlon,  Mathieu and Raffel,  Colin and von Werra,  Leandro and Wolf,  Thomas},
  title = {SmolLM2: When Smol Goes Big -- Data-Centric Training of a Small Language Model},
  publisher = {arXiv},
  year = {2025},
  copyright = {Creative Commons Attribution 4.0 International}
}

@misc{agriclip,
  doi = {10.48550/ARXIV.2410.01407},
  url = {https://arxiv.org/abs/2410.01407},
  author = {Nawaz,  Umair and Awais,  Muhammad and Gani,  Hanan and Naseer,  Muzammal and Khan,  Fahad and Khan,  Salman and Anwer,  Rao Muhammad},
  title = {AgriCLIP: Adapting CLIP for Agriculture and Livestock via Domain-Specialized Cross-Model Alignment},
  publisher = {arXiv},
  year = {2024},
  copyright = {Creative Commons Attribution 4.0 International}
}

@inproceedings{
crop,
title={Empowering and Assessing the Utility of Large Language Models in Crop Science},
author={Hang Zhang and Jiawei Sun and Renqi Chen and Wei Liu and Zhonghang Yuan and Xinzhe Zheng and Zhefan Wang and Zhiyuan Yang and Hang Yan and Han-Sen Zhong and Xiqing Wang and Wanli Ouyang and Fan Yang and Nanqing Dong},
booktitle={The Thirty-eight Conference on Neural Information Processing Systems Datasets and Benchmarks Track},
year={2024},
url={https://openreview.net/forum?id=hMj6jZ6JWU}
}

@misc{clip,
  doi = {10.48550/ARXIV.2103.00020},
  url = {https://arxiv.org/abs/2103.00020},
  author = {Radford,  Alec and Kim,  Jong Wook and Hallacy,  Chris and Ramesh,  Aditya and Goh,  Gabriel and Agarwal,  Sandhini and Sastry,  Girish and Askell,  Amanda and Mishkin,  Pamela and Clark,  Jack and Krueger,  Gretchen and Sutskever,  Ilya},
  title = {Learning Transferable Visual Models From Natural Language Supervision},
  publisher = {arXiv},
  year = {2021},
  copyright = {arXiv.org perpetual,  non-exclusive license}
}

@misc{SigLIP2,
  doi = {10.48550/ARXIV.2502.14786},
  url = {https://arxiv.org/abs/2502.14786},
  author = {Tschannen,  Michael and Gritsenko,  Alexey and Wang,  Xiao and Naeem,  Muhammad Ferjad and Alabdulmohsin,  Ibrahim and Parthasarathy,  Nikhil and Evans,  Talfan and Beyer,  Lucas and Xia,  Ye and Mustafa,  Basil and Hénaff,  Olivier and Harmsen,  Jeremiah and Steiner,  Andreas and Zhai,  Xiaohua},
  title = {SigLIP 2: Multilingual Vision-Language Encoders with Improved Semantic Understanding,  Localization,  and Dense Features},
  publisher = {arXiv},
  year = {2025},
  copyright = {Creative Commons Attribution 4.0 International}
}

@misc{PlantVillageVQA,
  doi = {10.48550/ARXIV.2508.17117},
  url = {https://arxiv.org/abs/2508.17117},
  author = {Sakib,  Syed Nazmus and Haque,  Nafiul and Hossain,  Mohammad Zabed and Arman,  Shifat E.},
  title = {PlantVillageVQA: A Visual Question Answering Dataset for Benchmarking Vision-Language Models in Plant Science},
  publisher = {arXiv},
  year = {2025},
  copyright = {Creative Commons Attribution Share Alike 4.0 International}
}

@inproceedings{rouge,
    title = "{ROUGE}: A Package for Automatic Evaluation of Summaries",
    author = "Lin, Chin-Yew",
    booktitle = "Text Summarization Branches Out",
    month = jul,
    year = "2004",
    address = "Barcelona, Spain",
    publisher = "Association for Computational Linguistics",
    url = "https://aclanthology.org/W04-1013/",
    pages = "74--81"
}

@misc{InternVL3,
  doi = {10.48550/ARXIV.2504.10479},
  url = {https://arxiv.org/abs/2504.10479},
  author = {Wang, Weiyun and Gao, Zhangwei and Gu, Lixin and Pu, Hengjun and Cui, Long and Wei, Xingguang and Liu, Zhaoyang and Jing, Linglin and Ye, Shenglong and Shao, Jie and others},
  title = {InternVL3: Exploring Advanced Training and Test-Time Recipes for Open-Source Multimodal Models},
  publisher = {arXiv},
  year = {2025},
  copyright = {Creative Commons Attribution 4.0 International}
}

@misc{gemma4-wang,
  doi = {10.48550/ARXIV.2604.07035},
  url = {https://arxiv.org/abs/2604.07035},
  author = {Manik,  Md Motaleb Hossen and Wang,  Ge},
  title = {Unified Deployment-Aware Evaluation of Open Reasoning Language Models},
  publisher = {arXiv},
  year = {2026},
  copyright = {Creative Commons Attribution 4.0 International}
}

@article{Gemini2.0,
  title = {Gender and content bias in Large Language Models: a case study on Google Gemini 2.0 Flash Experimental},
  volume = {8},
  ISSN = {2624-8212},
  url = {http://dx.doi.org/10.3389/frai.2025.1558696},
  DOI = {10.3389/frai.2025.1558696},
  journal = {Frontiers in Artificial Intelligence},
  publisher = {Frontiers Media SA},
  author = {Balestri,  Roberto},
  year = {2025},
  month = Mar 
}

@misc{Qwen2.5,
  doi = {10.48550/ARXIV.2412.15115},
  url = {https://arxiv.org/abs/2412.15115},
  author = {Yang,  An and Yang,  Baosong and Zhang,  Beichen and Hui,  Binyuan and Zheng,  Bo and Yu,  Bowen and Li,  Chengyuan and Liu,  Dayiheng and Huang,  Fei and Wei,  Haoran and Lin,  Huan and Yang,  Jian and Tu,  Jianhong and Zhang,  Jianwei and others},
  title = {Qwen2.5 Technical Report},
  publisher = {arXiv},
  year = {2024},
  copyright = {arXiv.org perpetual,  non-exclusive license}
}

@misc{Qwen3,
  doi = {10.48550/ARXIV.2505.09388},
  url = {https://arxiv.org/abs/2505.09388},
  author = {Yang,  An and Li,  Anfeng and Yang,  Baosong and Zhang,  Beichen and Hui,  Binyuan and Zheng,  Bo and Yu,  Bowen and Gao,  Chang and Huang,  Chengen and Lv,  Chenxu and Zheng,  Chujie and Liu,  Dayiheng and Zhou,  Fan and Huang,  Fei and Hu,  Feng and others},
  title = {Qwen3 Technical Report},
  publisher = {arXiv},
  year = {2025},
  copyright = {arXiv.org perpetual,  non-exclusive license}
}

@misc{LlaVA,
  doi = {10.48550/ARXIV.2304.08485},
  url = {https://arxiv.org/abs/2304.08485},
  author = {Liu,  Haotian and Li,  Chunyuan and Wu,  Qingyang and Lee,  Yong Jae},
  title = {Visual Instruction Tuning},
  publisher = {arXiv},
  year = {2023},
  copyright = {Creative Commons Attribution 4.0 International}
}

@misc{LFM2,
  doi = {10.48550/ARXIV.2511.23404},
  url = {https://arxiv.org/abs/2511.23404},
  author = {Amini,  Alexander and Banaszak,  Anna and Benoit,  Harold and B\"{o}\"{o}k,  Arthur and Dakhran,  Tarek and Duong,  Song and Eng,  Alfred and Fernandes,  Fernando and H\"{a}rk\"{o}nen,  Marc and Harrington,  Anne and Hasani,  Ramin and Karwa,  Saniya and Khrustalev,  Yuri and others},
  title = {LFM2 Technical Report},
  publisher = {arXiv},
  year = {2025},
  copyright = {Creative Commons Attribution 4.0 International}
}

@misc{BioCLIP,
  doi = {10.48550/ARXIV.2311.18803},
  url = {https://arxiv.org/abs/2311.18803},
  author = {Stevens,  Samuel and Wu,  Jiaman and Thompson,  Matthew J and Campolongo,  Elizabeth G and Song,  Chan Hee and Carlyn,  David Edward and Dong,  Li and Dahdul,  Wasila M and Stewart,  Charles and Berger-Wolf,  Tanya and Chao,  Wei-Lun and Su,  Yu},
  title = {BioCLIP: A Vision Foundation Model for the Tree of Life},
  publisher = {arXiv},
  year = {2023},
  copyright = {Creative Commons Attribution 4.0 International}
}

@misc{SCOLD,
  doi = {10.48550/ARXIV.2505.07019},
  url = {https://arxiv.org/abs/2505.07019},
  author = {Quoc,  Khang Nguyen and Thu,  Lan Le Thi and Quach,  Luyl-Da},
  title = {A Vision-Language Foundation Model for Leaf Disease Identification},
  publisher = {arXiv},
  year = {2025},
  copyright = {Creative Commons Attribution Share Alike 4.0 International}
}

@article{Jafar2024,
  title = {Revolutionizing agriculture with artificial intelligence: plant disease detection methods,  applications,  and their limitations},
  volume = {15},
  ISSN = {1664-462X},
  url = {http://dx.doi.org/10.3389/fpls.2024.1356260},
  DOI = {10.3389/fpls.2024.1356260},
  journal = {Frontiers in Plant Science},
  publisher = {Frontiers Media SA},
  author = {Jafar,  Abbas and Bibi,  Nabila and Naqvi,  Rizwan Ali and Sadeghi-Niaraki,  Abolghasem and Jeong,  Daesik},
  year = {2024},
  month = Mar 
}

@article{Khan2025,
  title = {A review on automated plant disease detection: motivation,  limitations,  challenges,  and recent advancements for future research},
  volume = {37},
  ISSN = {2213-1248},
  url = {http://dx.doi.org/10.1007/s44443-025-00040-3},
  DOI = {10.1007/s44443-025-00040-3},
  number = {3},
  journal = {Journal of King Saud University Computer and Information Sciences},
  publisher = {Springer Science and Business Media LLC},
  author = {Khan,  Sajid Ullah and Alsuhaibani,  Anas and Alabduljabbar,  Abdulrahman and Almarshad,  Fahdah and Altherwy,  Youssef N. and Akram,  Tallha},
  year = {2025},
  month = May 
}

@article{Sajitha2024,
  title = {A review on machine learning and deep learning image-based plant disease classification for industrial farming systems},
  volume = {38},
  ISSN = {2452-414X},
  url = {http://dx.doi.org/10.1016/j.jii.2024.100572},
  DOI = {10.1016/j.jii.2024.100572},
  journal = {Journal of Industrial Information Integration},
  publisher = {Elsevier BV},
  author = {Sajitha,  P. and Andrushia,  A. Diana and Anand,  N. and Naser,  M.Z.},
  year = {2024},
  month = Mar,
  pages = {100572}
}

@article{Quach2025,
  title = {XAI-BO: an architecture using Grad-CAM technique to evaluate Bayesian optimization algorithms on deep learning models},
  volume = {9},
  ISSN = {2475-1847},
  url = {http://dx.doi.org/10.1080/24751839.2024.2447191},
  DOI = {10.1080/24751839.2024.2447191},
  number = {3},
  journal = {Journal of Information and Telecommunication},
  publisher = {Informa UK Limited},
  author = {Quach,  Luyl-Da and Quoc Khang,  Nguyen and Thai-Nghe,  Nguyen and Nguyen,  Chi-Ngon},
  year = {2025},
  month = Jan,
  pages = {335–356}
}

@article{SenthilPandi2022,
  title = {Rice plant disease classification using dilated convolutional neural network with global average pooling},
  volume = {474},
  ISSN = {0304-3800},
  url = {http://dx.doi.org/10.1016/j.ecolmodel.2022.110166},
  DOI = {10.1016/j.ecolmodel.2022.110166},
  journal = {Ecological Modelling},
  publisher = {Elsevier BV},
  author = {Senthil Pandi,  S and Senthilselvi,  A and Gitanjali,  J and ArivuSelvan,  K and Gopal,  Jagadeesh and Vellingiri,  J},
  year = {2022},
  month = Dec,
  pages = {110166}
}

@article{Zhu2025,
  title = {Harnessing large vision and language models in agriculture: a review},
  volume = {16},
  ISSN = {1664-462X},
  url = {http://dx.doi.org/10.3389/fpls.2025.1579355},
  DOI = {10.3389/fpls.2025.1579355},
  journal = {Frontiers in Plant Science},
  publisher = {Frontiers Media SA},
  author = {Zhu,  Hongyan and Qin,  Shuai and Su,  Min and Lin,  Chengzhi and Li,  Anjie and Gao,  Junfeng},
  year = {2025},
  month = Sept 
}

@article{Tzachor2023,
  title = {Large language models and agricultural extension services},
  volume = {4},
  ISSN = {2662-1355},
  url = {http://dx.doi.org/10.1038/s43016-023-00867-x},
  DOI = {10.1038/s43016-023-00867-x},
  number = {11},
  journal = {Nature Food},
  publisher = {Springer Science and Business Media LLC},
  author = {Tzachor,  A. and Devare,  M. and Richards,  C. and Pypers,  P. and Ghosh,  A. and Koo,  J. and Johal,  S. and King,  B.},
  year = {2023},
  month = Nov,
  pages = {941–948}
}

@misc{Late_Blight,
  title = {Late Blight of Tomato},
  ISBN = {9781118728475},
  url = {http://dx.doi.org/10.1002/9781118728475.ch13},
  DOI = {10.1002/9781118728475.ch13},
  journal = {Translational Genomics for Crop Breeding},
  publisher = {Wiley},
  author = {Nowicki,  Marcin and Kozik,  Elżbieta U. and Foolad,  Majid R.},
  year = {2013},
  month = Oct,
  pages = {241–265}
}

@misc{Bacterial_spot,
  title = {Xanthomonas vesicatoria (bacterial spot of tomato and pepper)},
  url = {http://dx.doi.org/10.1079/cabicompendium.56981},
  DOI = {10.1079/cabicompendium.56981},
  journal = {CABI Compendium},
  publisher = {CABI Publishing},
  author = {Osdaghi,  Ebrahim},
  year = {2020},
  month = Dec 
}

@article{Fusarium,
  title = {Fusarium oxysporum f. sp. lycopersici causal agent of vascular wilt disease of tomato: Biology to diversity-- A review},
  volume = {26},
  ISSN = {1319-562X},
  url = {http://dx.doi.org/10.1016/j.sjbs.2019.06.002},
  DOI = {10.1016/j.sjbs.2019.06.002},
  number = {7},
  journal = {Saudi Journal of Biological Sciences},
  publisher = {Elsevier BV},
  author = {Srinivas,  C. and Nirmala Devi,  D. and Narasimha Murthy,  K. and Mohan,  Chakrabhavi Dhananjaya and Lakshmeesha,  T.R. and Singh,  BhimPratap and Kalagatur,  Naveen Kumar and Niranjana,  S.R. and Hashem,  Abeer and Alqarawi,  Abdulaziz A. and Tabassum,  Baby and Abd\_Allah,  Elsayed Fathi and Chandra Nayaka,  S. and Srivastava,  Rakesh K.},
  year = {2019},
  month = Nov,
  pages = {1315--1324}
}

@article{Leaf_mold,
  title = {Understanding the mechanisms of resistance to tomato leaf mold: A review},
  volume = {8},
  ISSN = {2468-0141},
  url = {http://dx.doi.org/10.1016/j.hpj.2022.04.008},
  DOI = {10.1016/j.hpj.2022.04.008},
  number = {6},
  journal = {Horticultural Plant Journal},
  publisher = {Elsevier BV},
  author = {Zhao,  Tingting and Pei,  Tong and Jiang,  Jingbin and Yang,  Huanhuan and Zhang,  He and Li,  Jingfu and Xu,  Xiangyang},
  year = {2022},
  month = Nov,
  pages = {667–675}
}

@article{Verticillium_wilt,
  title = {Studies on the nature of resistance in tomato plants to Verticillium albo‐atrum},
  volume = {62},
  ISSN = {1744-7348},
  url = {http://dx.doi.org/10.1111/j.1744-7348.1968.tb02827.x},
  DOI = {10.1111/j.1744-7348.1968.tb02827.x},
  number = {2},
  journal = {Annals of Applied Biology},
  publisher = {Wiley},
  author = {SINHA,  A. K. and WOOD,  R. K. S.},
  year = {1968},
  month = Oct,
  pages = {319–327}
}

@article{Bacterial_Wilt,
title = {Impact of Soil Moisture Regimes on Wilt Disease in Tomatoes: Current Understanding},
ISBN = {9780128130667},
url = {http://dx.doi.org/10.1016/B978-0-12-813066-7.00005-X},
DOI = {10.1016/b978-0-12-813066-7.00005-x},
booktitle = {Biochemical,  Physiological and Molecular Avenues for Combating Abiotic Stress Tolerance in Plants},
publisher = {Elsevier},
author = {Gupta,  Aarti and Kamalachandran,  Dharanipathi and Longchar,  Bendangchuchang and Senthil-Kumar,  Muthappa},
year = {2018},
pages = {73–82}
}

@article{Early_Blight,
  title = {Alternaria Solani on Tomato},
  volume = {154},
  ISSN = {1476-4687},
  url = {http://dx.doi.org/10.1038/154642a0},
  DOI = {10.1038/154642a0},
  number = {3916},
  journal = {Nature},
  publisher = {Springer Science and Business Media LLC},
  author = {GLASSCOCK,  W. H. and WARE,  W. M.},
  year = {1944},
  month = Nov,
  pages = {642–642}
}

@article{septoria_leaf,
  title = {Septoria Leaf Spot of Tomatoes: Historical Insights,  Present Challenges,  and Future Prospects},
  volume = {10},
  ISSN = {2311-7524},
  url = {http://dx.doi.org/10.3390/horticulturae10121299},
  DOI = {10.3390/horticulturae10121299},
  number = {12},
  journal = {Horticulturae},
  publisher = {MDPI AG},
  author = {Pandey,  Anju and Paudel,  Rajan and Adhikari,  Tika B. and Panthee,  Dilip R. and Louws,  Frank J.},
  year = {2024},
  month = Dec,
  pages = {1299}
}

@article{target_spot,
  title = {AN OVERVIEW OF TARGET SPOT OF TOMATO CAUSED BY CORYNESPORA CASSIICOLA},
  ISSN = {2406-6168},
  url = {http://dx.doi.org/10.17660/ActaHortic.2009.808.1},
  DOI = {10.17660/actahortic.2009.808.1},
  number = {808},
  journal = {Acta Horticulturae},
  publisher = {International Society for Horticultural Science (ISHS)},
  author = {Schlub,  R.L. and Smith,  L.J. and Datnoff,  L.E. and Pernezny,  K.},
  year = {2009},
  month = Jan,
  pages = {25–28}
}

@article{yellow_leaf_curl,
  title = {A worldwide survey of tomato yellow leaf curl viruses},
  volume = {142},
  ISSN = {1432-8798},
  url = {http://dx.doi.org/10.1007/s007050050168},
  DOI = {10.1007/s007050050168},
  number = {7},
  journal = {Archives of Virology},
  publisher = {Springer Science and Business Media LLC},
  author = {Czosnek,  H. and Laterrot,  H.},
  year = {1997},
  month = July,
  pages = {1391–1406}
}

@inbook{mosaic_virus,
  title = {Tomato Mosaic Virus},
  ISBN = {9783031818844},
  url = {http://dx.doi.org/10.1007/978-3-031-81884-4_21},
  DOI = {10.1007/978-3-031-81884-4_21},
  booktitle = {Compendium of Phytopathogenic Microbes in Agro-Ecology },
  publisher = {Springer Nature Switzerland},
  author = {Naik,  Ami and Chaudhary,  Ankit},
  year = {2025},
  pages = {329–347}
}

@article{spider,
  title = {The genome of Tetranychus urticae reveals herbivorous pest adaptations},
  volume = {479},
  ISSN = {1476-4687},
  url = {http://dx.doi.org/10.1038/nature10640},
  DOI = {10.1038/nature10640},
  number = {7374},
  journal = {Nature},
  publisher = {Springer Science and Business Media LLC},
  author = {Grbić,  Miodrag and Van Leeuwen,  Thomas and Clark,  Richard M. and Rombauts,  Stephane and Rouzé,  Pierre and Grbić,  Vojislava and Osborne,  Edward J. and others},
  year = {2011},
  month = Nov,
  pages = {487–492}
}
\clearpage
\setcounter{page}{1}

\appendix
\begin{center}
{\Large\bfseries TomaMMU: A Comprehensive Multimodal
Understanding Benchmark for Tomato Leaf
Diseases}\\[0.65em]

{\large\bfseries -- Supplementary Material --}
\end{center}

\setcounter{table}{0}
\renewcommand{\thetable}{S\arabic{table}}

\section{Reference Papers for Agronomic Reasoning}
\label{sec:agronomic_references}

\begin{table}[htbp]
\centering
\caption{Reference papers utilized for the agronomic reasoning of tomato diseases.}
\label{tab:disease_papers}
\renewcommand{\arraystretch}{1.3}
\begin{tabular}{l p{8.5cm}}
\toprule
Disease Name & Reference Paper \\
\midrule
Bacterial Spot & Xanthomonas vesicatoria (bacterial spot of tomato and pepper) \cite{bacterial_spot} \\

Bacterial Wilt & Impact of Soil Moisture Regimes on Wilt Disease in Tomatoes: Current Understanding \cite{Bacterial_Wilt} \\

Early Blight & Alternaria Solani on Tomato \cite{Early_Blight} \\

Late Blight & Late Blight of Tomato \cite{Late_Blight}\\

Fusarium Wilt & Fusarium oxysporum f. sp. lycopersici causal agent of vascular wilt disease of tomato: Biology to diversity– A review \cite{Fusarium} \\

Verticillium Wilt &  Studies on the nature of resistance in tomato plants to Verticillium albo‐atrum
\cite{Verticillium_wilt} \\

Leaf Mold & Understanding the mechanisms of resistance to tomato leaf mold: A review
\cite{Leaf_mold}\\

Septoria Leaf Spot & Septoria Leaf Spot of Tomatoes: Historical Insights,  Present Challenges,  and Future Prospects
\cite{septoria_leaf}\\

Target Spot & An overview of target spot of tomato caused by corynespora cassiicola 
\cite{target_spot}\\

Yellow Leaf Curl Virus & A worldwide survey of tomato yellow leaf curl viruses 
\cite{yellow_leaf_curl}\\

Mosaic Virus & Tomato mosaic virus \cite{mosaic_virus}\\

Spider Mites & The genome of Tetranychus urticae reveals herbivorous pest adaptations
\cite{spider}\\
\bottomrule
\end{tabular}
\end{table}

\section{Prompt Template for GPT Evaluation}
\label{sec:prompt_template}
\FloatBarrier
To evaluate the performance of our candidate models and compute the GPT score, we utilize the structured evaluation prompt presented in Table \ref{tab:prompt_template}.

\begin{table}[t!]
\centering
\caption{Prompt template utilized for the automated evaluation of prediction accuracy.}
\label{tab:prompt_template}
\renewcommand{\arraystretch}{1.3}
\begin{tabular}{p{0.95\textwidth}}
\toprule
\textbf{System Prompt \& Role} \\ \midrule
You are an expert multimodal AI evaluator. Your task is to evaluate the accuracy of a candidate answer to a visual question, based on the provided ground truth answer set. \\
\\
\textbf{[Input Data]} \\
$\bullet$ \textbf{Ground Truth Reference Answers:} \{ground\_truth\_answers\} \\
$\bullet$ \textbf{Prediction Answer:} \{prediction\_answer\} \\
\\
\textbf{[Evaluation Criteria]} \\
Evaluate the candidate answer based on the following scale (1 to 5): \\
\quad\textbf{1:} The answer is completely incorrect, hallucinatory, or irrelevant to the question. \\
\quad\textbf{2:} The answer is partially relevant but contains major inaccuracies or misses the core question. \\
\quad\textbf{3:} The answer is generally correct but lacks specific details, or contains minor factual errors. \\
\quad\textbf{4:} The answer is correct, directly addresses the question, and is semantically equivalent to the ground truth. \\
\quad\textbf{5:} The answer is exceptionally accurate, clear, and demonstrates a strong understanding of the visual and textual context. \\
\\
\textbf{[Output Format]} \\
Provide your evaluation and give me the average score of the model. \\
\bottomrule
\end{tabular}
\end{table}
\FloatBarrier
\end{document}